\documentclass{article}
\usepackage{spconf,amsmath,graphicx}
\usepackage{booktabs}
\usepackage{float}

\title{ACTION STATE UPDATE APPROACH TO DIALOGUE MANAGEMENT}

\name{Svetlana Stoyanchev, Simon Keizer and Rama Doddipatla \thanks{$\copyright$ 2021 IEEE.  Personal use of this material is permitted.  Permission from IEEE must be obtained for all other uses, in any current or future media, including reprinting/republishing this material for advertising or promotional purposes, creating new collective works, for resale or redistribution to servers or lists, or reuse of any copyrighted component of this work in other works.}}
\address{Toshiba Cambridge Research Laboratory, Cambridge, UK}

\begin{document}
\ninept
\maketitle
\begin{abstract}
Utterance interpretation is one of the main functions of a dialogue manager, which is the key component of a dialogue system. We propose the action state update approach (ASU) for utterance interpretation, featuring a statistically trained binary classifier used to detect dialogue state update actions in the text of a user utterance. Our goal is to interpret referring expressions in user input without a domain-specific natural language understanding component. For training the model, we use active learning to automatically select simulated training examples. With both user-simulated and interactive human evaluations, we show that the ASU approach successfully interprets user utterances in a dialogue system, including those with referring expressions.
\end{abstract}

\begin{figure*}[htb]
    \centering
    \includegraphics[width=0.9\textwidth]{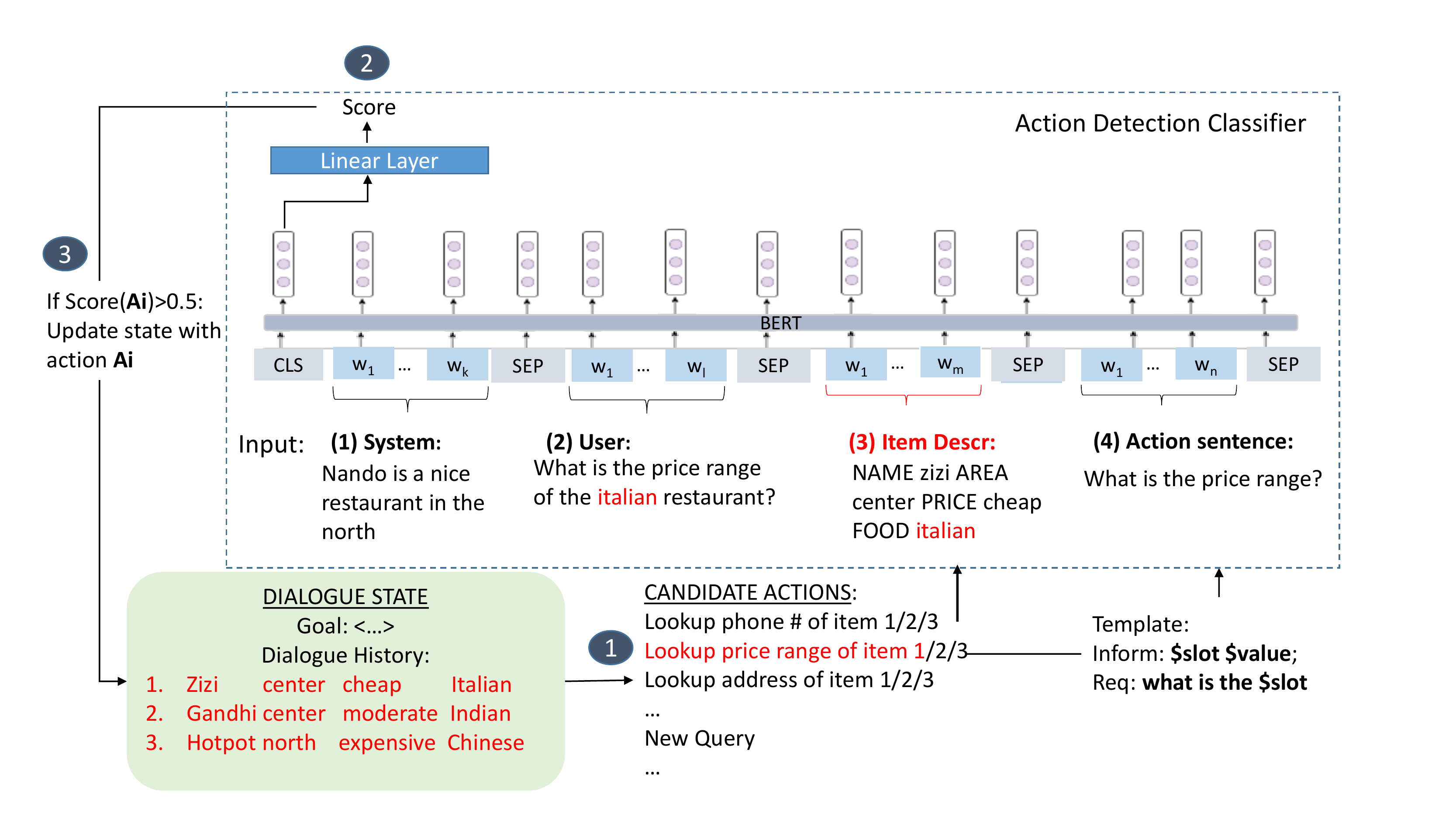}
    \caption{Action State Update Model: 1) infer candidate actions from the state; 2) score each action 3) update the state}
    \label{fig:model}
\end{figure*}

\section{Introduction}
\label{sec:intro}

Task-oriented dialogue systems are natural language interfaces for tasks, such as information search,  customer support,  e-commerce, physical environment  control, and human-robot interaction.
Natural language is a universal communication interface that does not require users to learn a set of task-specific commands.
A spoken interface allows the user to communicate by speaking, and a chat interface by typing.
Correct interpretation of user input can be challenging for automatic dialogue systems which lack the grammatical and common sense knowledge that allows people to effortlessly interpret a wide variety of natural input.

\begin{table}[ht]
\begin{center}
\begin{tabular}{ rlp{6cm} }
\toprule
1&Usr: & {\it I am looking for a cheap {\bf Italian} restaurant.}   \\
2&Sys: & {\bf Zizzi Cambridge} is a nice place in the center. \\
3&Usr: & {\it How about Indian?}\\
4&Sys: & Nando is a cheap Indian place you might like.\\
5&Usr: & {\it What is the address of {\bf the Italian place}?}\\
6&Sys: &  The address of Zizzi Cambridge is ...\\
\bottomrule
\end{tabular}
\caption{Dialogue example.}
\label{tbl:dialog_example}
\end{center}
\end{table}

Referring expressions allow a speaker to verbally identify entities in dialogue context among multiple candidates. For example, in the fifth turn of the dialogue in Table~\ref{tbl:dialog_example}, the user asks for the address of a restaurant presented by the system three turns earlier (Zizzi) and following a presentation of another restaurant (Nando). The user identifies the target restaurant with the referring expression {\it `the Italian place'}.
 The semantic representation for this question has two arguments: the {\em target}, describing the information requested, and the {\em reference}, describing the entity referred to.

Referring expressions pose additional challenges to dialogue systems. To interpret a user input with a referring expression, the system not only has to identify the intent and entities but also the specific item in context that the user is referring to. The ability to interpret and respond to user requests with referring expressions is a functionality required in real-world intelligent interfaces.

Reference resolution in text and dialogue has been extensively addressed in past research~\cite{DBLP:series/tanlp/2016anaphora,Maptask,monroe2017,5700875,el-asri-etal-2017-frames}.
While it is recognized as a key part of discourse interpretation, reference resolution has not been the focus of the recent dialogue challenges and referring expressions are not induced or annotated in most of the recent task-oriented dialogue corpora~\cite{henderson-thomson-williams:2014:W14-43,budzianowski-etal-2018-multiwoz,schema-guided}.
In a dialogue system with a traditional pipeline architecture, a Natural Language Understanding component (NLU) assigns semantics to a user utterance~\cite{williams2015fast,mehri2020dialoglue}.
In the Information State Update (ISU) approach, the system maintains a representation of the dialogue context, called the `Information State', using a set of update rules.  These update rules can be triggered by `dialogue moves', which are semantic representations of utterances, generated by a domain-specific NLU~\cite{traum_information_2003}.  As the Information State can store previously discussed entities, an ISU system is capable of interpreting the referring expressions in the user input.  However, to interpret referring expressions these systems would require a task-specific NLU which is costly and time-consuming to train~\cite{chen:etal:18}.

In the POMDP approach, the aim is to create a system that is more robust to noise by maintaining a probability distribution over dialogue states.  To make this tractable, the dialogue states have to be compact and therefore typically only model the user goal and very limited dialogue history information~\cite{YOUNG2010150}.
While initial models still relied on ISU style rule-based state updating, later approaches used trainable statistical dialogue state tracking (DST)~\cite{williams-2012-belief}.
Furthermore, jointly trained NLU, DST, and policy components have been shown to improve system performance~\cite{7953246,wenN2N17}.
Recently, neural approaches have enabled text-based state tracking, avoiding the need for a separate NLU component~\cite{henderson-etal-2014-word,rastogi-etal-2018-multi,mrksic-etal-2017-neural}.  However, these approaches do not handle referring expressions.

We propose the {\it action state update} approach (ASU), which interprets referring expressions without a domain-specific NLU and combines the advantages of flexible state representation, statistical state updating, and declarative representation~\cite{AAAIW1817387,stoyanchev-lison-bangalore:2016:SIGDIAL}.
ASU uses a binary classifier model to detect the state update operations (called {\em actions}) in a user utterance. The action set is dynamically generated based on the current state. Hence we use a binary classifier with a multi-pass approach, where the classifier makes a prediction for each action in a turn instead of using a multi-class classification approach typically used for state tracking in slot-filling systems.
The proposed approach is intended for both spoken and chat natural language interfaces.  In this work, we evaluate it with a chat interface where the user interacts with a dialogue system by typing phrases or sentences.
Using active learning, the proposed ASU model achieves 98.3\% state update accuracy in text-based simulation. We also evaluate the model with the recruited subjects using the chat interface.

The contributions of this work are:
\begin{itemize}
 \item We introduce a novel approach to state updating that does not require NLU and interprets referring expressions.
 \item We show that the proposed (binary) model trained with the distractors generated using active learning significantly improves the model performance.
\end{itemize}

\section{Method}
\label{sec:method}

\begin{table*}[th]
\begin{center}
\begin{tabular}{ |l|l|l|l|l| }
\hline
Req.& Ref. &Generic & Template  & Simulated user request \\
slot & slot & request& & \\\hline\hline
food & name &  - & What type of food does \$name serve? &  What type of food does zizzi serve?  \\\hline
price & area & price range& for the restaurant  in the \$area  & price range for the place in the center \\\hline
area & price &  area & of the \$price place & area of the cheap place\\\hline
area & food &  what's the area & for the \$food place & what's the area for the italian place\\\hline
\end{tabular}
\caption{Generated requests with referring expressions for sampled item (zizzi, cheap, italian, center).}
\label{tbl:simreq}
\end{center}
\end{table*}

\begin{table}[th]
\begin{center}
\begin{tabular}{ |l|p{3.7cm}|c| }
\hline
Model & Generation method & train/dev data size  \\
& & (\% positive) \\\hline
$init$ & from DSTC2 & 72 / 24K (15\%)  \\\hline
~~+$ext_H$ & expand w. heuristics & 137 / 43K (12\%) \\\hline
~~+$ext_A$ & expand w. active learning & 101 / 31K (16\%)\\\hline
\end{tabular}
\caption{Training datasets for the experimental models.}
\label{tbl:data}
\end{center}
\end{table}

We implement a dialogue system for the Cambridge restaurant search domain~\cite{henderson-thomson-williams:2014:W14-43}.
Similarly to other slot-filling systems, our system allows a user to find a restaurant matching specified {\it area, price range,} or {\it food type}  (the {\it informable} slots in the domain definition). A user may change the search criteria and explore alternative restaurants. In contrast to the other slot-filling systems,  our system allows a user to request a {\it phone number, address, post code,  area, price range,} or {\it food type} about any of the previously discussed restaurants.
The system has three modules: 1) an {\it action state update} component is the focus of this work (see Section~\ref{sec:AD}), 2) a system move selection component, and 3) a template-based natural language generator.\footnote{A rule-based generic NLU handles social interaction but is outside of the scope for the current experiment.} The system move selection policy is trained using reinforcement learning in interaction with an agenda-based user simulator~\cite{schatzmann-etal-2007-agenda}.

\subsection{Action State Update}
\label{sec:AD}

In our approach, a state update is seen as a set of operations, or {\em actions}. Each {\em action} changes a value in the dialogue state, which stores the system beliefs about the user goal and dialogue history, including previously discussed items.  For example, a state update {\it action} for the utterance {\it `I am interested in Italian food'} updates the user goal with {\it food=Italian}.  A state update {\it action} for the utterance {\it `What area is the Italian restaurant in?'} switches {\it on} a request bit for the {\it area} field of the entity matching the property {\it food=Italian}.
{\it Action detection} is the task of identifying which state modifying actions are intended by the user in a given context. In our approach, {\it actions}, which are instructions for the state modification, are detected without a semantic parse of the utterance.

On each user turn, the ASU model (Figure~\ref{fig:model}): 1) infers the candidate actions from the dialogue state; 2) computes a relevance $score \in [0,1]$ for each candidate action; and 3) updates the state with the most likely {\it actions}.
To allow the user to ask a question about any of the discussed items, a candidate $request$ action is generated for each of the {\it requestable slots} for each item stored in the dialogue history.  For example, if the dialogue history contains three restaurants, 18 request candidate actions are generated (6 requestable slots x 3 items). Changing the user goal, in contrast, is a context-independent action.  Given the domain ontology, the model classifies the same number of the {\it goal changing actions} in each turn, corresponding to the (informable) slot-value pairs.\footnote{In Cambridge restaurants domain has 102 values for the {\it food type, area, } and {\it pricerange} slots.}

At run-time, for each turn the model makes N classification decisions where N is the number of {\bf dynamically generated} candidate actions in that turn. Only the actions with a relevance score above 0.5 are executed, i.e., update the dialogue state by either changing the goal (a slot value) or setting a request bit on one of the items in the dialogue history. During the update, we apply the following heuristics: 1) if multiple actions for a slot are predicted, we use the one with the highest $score$; 2) if multiple request actions receive $score>0.5$, we update the request bit for the most recently mentioned item only.

Following the success of transfer learning with pretrained Transformer models on various NLP tasks,
we adopt the pre-trained BERT model in the {\it action detection} task.  
The input to the model is a word sequence, consisting of: 1) a sequence of lexicalized dialogue acts\footnote{Using lexicalized dialogue acts achieved higher performance than using the system utterance.}, 2) a user utterance, 3) an  {\bf item description}, and 4) a template-generated {\bf action sentence}.
An item description is a string generated from the action. For item-independent actions (goal changes), the item description is empty; for item-dependent actions (information requests), it corresponds to the description of the requested item. The description corresponding to the action {\it request address of the first item} for the state in Figure~\ref{fig:model} is {\it `NAME zizi AREA center PRICE cheap FOOD italian'}.
By including an item description in the model input, the attention mechanism of the transformer model learns to detect whether an action can be inferred from a user utterance in a given context.
The presence of the item description, the dynamic generation of candidate actions, and the method of data generation allow the model to interpret referring expressions, which differentiates our approach from previous work that also uses a binary BERT model for state tracking~\cite{simplebert}.

\subsection{Data generation}
\label{sec:data}

The AD model is trained with positive and negative examples:
\begin{center}
\begin{tabular}{ c }
 $<sys, usr, action\rightarrow(itemdescr, actionsent)>: 0/1$ \\
\end{tabular}
\end{center}

In the positive examples (labeled 1), the {\it action} is intended by the user and in the negative examples (labeled 0), it is not.

Since {\it action} is an instruction on the current state, e.g. {\it `request pricerange of the first item'}, the {\it item description} and {\it action sentence} inputs to the model are inferred from the {\it action} and the state.
We generate three datasets for training the ASU action detection classifier summarized in Table~\ref{tbl:data}.

The $baseline$ dataset is generated from the training split of the DSTC2 corpus. For each turn, we generate a positive example for each action {\it intended} by the user. The {\it intended} actions are inferred from the manual NL annotation. To generate the negative examples (distractors), we considered using all valid {\it unintended} actions (slot-value pairs). However, this creates a highly skewed dataset when the number of actions is large. Instead, for each positive example, we sample {\it unintended} actions using frequency and similarity heuristics to select more relevant distractors\footnote{The number of negative examples (5) was selected heuristically.}.
By the design of the task, the DSTC2 dataset does not contain referring expressions in user turns. All user requests are generic and refer to the last presented item (e.g., {\it What is the phone number?}). Hence, the model trained on the $baseline$ dataset can only understand references to the last presented item.

The $ext_H$ extends the $baseline$ dataset with the automatically generated utterances with referring expressions. A user may ask a question about any of the {\it requestable} slots and refer to any of the {\it informable} slots.
We generate 10K / 3K requests with referring expressions for train / dev dataset for all combinations of  {\it requestable} and {\it informable} slots by randomly sampling a request utterance without a referring expression for the request slot from DSTC2 dataset and concatenating it with a template-generated referring expression for the reference slots (see Table~\ref{tbl:simreq}).

The key idea of {\it active learning} is to allow an algorithm to select the training examples to learn from~\cite{settles2009active}. The $ext_A$ dataset is generated by automatically selecting the most challenging distractors from simulated dialogues.
We have extended our user simulator to explore multiple venues by repeatedly changing the goal constraints and then request slots for venues that were offered earlier in the dialogue.  In addition, we created templates for generating utterances with referring expressions for this new behaviour, resulting in a hybrid retrieval/template based model for generating simulated user utterances.
We first run the simulation with the ASU module using the classifier trained on the $baseline$ dataset for 5000 dialogues.
From the simulated user intents, we infer the {\it `intended'} user actions and automatically label the new training examples.
 Each {\it `intended'} action for which the $baseline$ model predicted a relevance $score<T_1$ is used as a positive example.
The top M {\it `unintended'} actions with the highest relevance $score>T_2$ are used as a negative example.\footnote{We set $T_1=.99$, $T_2=0.5$, and $M=2$ in this experiment.}
All generated utterances with referring expressions are also used as positive examples, even if they were correctly classified with the model trained on the $baseline$ dataset.

\section{Experiments}
\label{sec:experiments}

\begin{table*}[htp]
\begin{center}
\begin{tabular}{ |l|l||c|c||c|c|c||c|c| }
\hline
ASU train set  & State Update & \multicolumn{2}{|c|}{Dialogue:} & \multicolumn{5}{|c|}{Turn: State Update Accuracy }  \\
               & method for& average & success              & all&\multicolumn{2}{|c|}{ with user {\it inform} act}&\multicolumn{2}{|c|}{ with user {\it request} act} \\
               &  Policy training & length       & rate    & accuracy& \# per dialog & accuracy & \# per dialog & accuracy (std)  \\\hline\hline
$baseline$  & DA & 10.06 & 43.9\%  & 50.0\%   & 4.6 & 58.6\%  & 3.3 &  30.9\% \\\hline\hline
$exp_H$ & DA &  9.17   & 91.1\% & 75.1\%  	& 3.9 & 79.0\%  & 2.4 &	50.0\%  \\\hline
$exp_H$ & ASU w. $exp_H$ & 9.97 & 92.0\% & 74.7\%   & 4.1 & 75.9\% & 2.2 & 54.3\%\\\hline
$exp_A$ & DA &  8.15 & {\bf 99.5\%}  & 98.1\%   	& {\bf 3.7} & {\bf 98.8\%}  & {\bf 1.5} & 94.0\% \\\hline
$exp_A$ &  ASU w. $exp_A$ &   {\bf 8.02} & 99.4\% &  {\bf 98.3\%}   	& {\bf 3.7} &  98.6\%  & {\bf 1.5} &	{\bf 95.4\%} \\\hline\hline
GOLD & DA & 7.93 & 99.8\%   &  100\%           &3.7 & 100\%          & 1.2 & 100\%  \\\hline
\end{tabular}
\vspace{-1ex}
\caption{Evaluation with a user simulator. The top experimental scores are {\bf highlighted}.}
\label{tbl:resultsSim}
\end{center}
\end{table*}

\begin{table}[htp]
\begin{center}
\begin{tabular}{lll}
\toprule
Average \# turns (std. dev) per user & \multicolumn{2}{r}{60.9 (16.0)}  \\
Average \% turns (std. dev) marked as error & \multicolumn{2}{r}{15\% (10.0\%)} \\
\midrule
\multicolumn{2}{l}{The system understood me well} & 4.4 \\
\multicolumn{2}{l}{The systems' responses were appropriate} & 4.3 \\
\multicolumn{2}{l}{I was able to retrieve the information about the venues} & 4.6 \\
\multicolumn{2}{l}{The system understood my references to the venues} & 4.8 \\
\multicolumn{2}{l}{I would recommend this system to my friend} & 3.9 \\
\multicolumn{2}{l}{How many of the 5 tasks were you able to complete?} & 3.6 \\
\bottomrule
\end{tabular}
\end{center}
\vspace{-3ex}
\caption{Human evaluation results.}
\label{tab:humRes}
\vspace{-3ex}
\end{table}

First, we evaluate the ASU approach trained with the baseline model on the test subset of the DSTC2 corpus, i.e., without referring expressions.
Using the manual transcript of the user input, the model correctly identifies 96\% of user informs and 99\% of user requests (average goal and request accuracy as computed by the official DSTC2 evaluation script).
Identifying requests is relatively simple when they do not refer to a specific item in the dialogue context.

\subsection{Evaluation with the simulated user}

Next, we evaluate the proposed approach on simulated dialogues with referring expressions in user requests. We run the simulation with the proposed {\it action state update} component trained on the $baseline$, $exp_H$, and $exp_A$ datasets (Table~\ref{tbl:resultsSim}). As an upper bound (GOLD) condition, we run the simulation with the correct actions inferred from the simulated dialogue acts.
The policy model is trained with the agenda-based simulation using dialogue acts (DA) as input and 25\% dialogue act confusion rate.
For the models trained on $exp_H$ and $exp_A$, we also train a policy with simulated user utterances, rather than dialogue act hypotheses, as input. In this condition, the policy may learn to overcome state update errors made by the ASU model.

We simulate 5000 dialogues for each experimental condition and report the statistics computed for the dialogues and individual turns.
The dialogue success rate is the proportion of the simulated dialogues where the system offered a venue matching the simulated user's goal constraints (possibly after a number of goal changes), and provided the additional information requested by the simulated user. The state update accuracy is computed as the average accuracy across: a) {\it all} turns, b) turns annotated as {\it inform} only, and c)  turns annotated as {\it request} only.

The simulated user behaviour is affected by the state update model. The average length of a simulated dialogue ranges between 7.93 for the GOLD condition and 10.06 for the $baseline$. The lower state update accuracy leads to longer dialogues because when the system fails to respond correctly, the simulated user repeats or rephrases the request increasing the dialogue length. The baseline condition achieves only 43.9\% dialogue success and 50.0\% state update accuracy on all user turns.
In the $exp_H$ DA condition, the dialogue success and the overall accuracy increase to 91.1\%  and 75.1\% with an accuracy of 79.0\% on {\it informs} but only 50.0\% on {\it requests}.
With the active learning approach ($exp_A$ DA), the dialogue success and the overall accuracy increase to 99.5\% and 98.1\% with an accuracy of 98.8\% on {\it informs} and 94.0\% on {\it requests}.
Using a matched policy affects the performance for both $exp_H$ and $exp_A$ models, increasing the accuracy on requests by 4.3 and 1.4 absolute \% points.  However, using the policy trained with the $exp_H$ model decreases the accuracy on user {\it inform} acts by 3.1 \% points and increases the dialogue length. Our results show that the action state update approach is effective in combination with active learning.

\subsection{Human evaluation}
\label{sec:human}

In order to test the proposed action detection model with real users, a preliminary user study was carried out.  As mentioned in Section~\ref{sec:method}, the text-based system consists of the proposed dialogue state tracker using the $exp_A$ action detection model, a dialogue policy trained with the text-based user simulator, and a template-based natural language generator.  Subjects were recruited and asked to carry out 5 tasks involving restaurant information navigation.  In each task, a subject was given an initial set of constraints (e.g., {\it food type: Chinese, price range: cheap}) and asked to get a suitable recommendation from the system.  They then continue their conversation to get 2 alternative recommendations by changing the constraints, obtaining 3 recommended venues in total.
Finally, they were asked to get additional information such as the phone number or the address for two of these venues.
Subjects were also asked to indicate when they felt a system response was incorrect, by entering $<${\it error}$>$.  After completing all 5 tasks, they filled out a questionnaire, consisting of 5 statements to score on a 6 point Likert scale, ranging from `strongly disagree' to `strongly agree', and a question asking how many tasks were successfully completed (see Table~\ref{tab:humRes}).

Each user entered 60.9 turns on average and marked 15\% of them as errors.
The questionnaire results indicate that the system understood their references to the venues (average score 4.8).
Half of the users indicated that they completed all five tasks and only one of the users felt that the system did not understand them well.  High standard deviation across users indicates high variability in user experience and possibly expectation of the system.
While there is room for improvement, the human evaluation shows that the proposed model can be used in an interactive dialogue system.

\section{Conclusions}
\label{sec:concl}
In this work, we introduced a novel approach for updating the dialogue state and showed that it can successfully interpret user utterances, including the requests with the referring expressions. We trained the experimental models by extending the initial Cambridge restaurants dataset with the simulated requests containing referring expressions and sampled distractors.
The model trained on the dataset where the distractors were sampled using the active learning approach, achieved the best performance  despite the smaller size of its training sets. The human evaluation of this model showed that the approach can be used in an dialogue system with real users. In future work, we will collect a larger dataset of human-computer dialogues with referring expressions and apply active learning to improve system performance.

\bibliographystyle{IEEEbib}
\bibliography{refs}

\end{document}